\documentclass[11pt,a4paper]{article}
\usepackage[hyperref]{naaclhlt2019}
\usepackage{times}
\usepackage{latexsym}
\usepackage{multirow}
\usepackage{gb4e}
\usepackage{lipsum}
\noautomath
\usepackage{graphicx} 
\usepackage{tikz-dependency}
\usepackage[export]{adjustbox} 

\usepackage{url}
\usepackage[belowskip=4pt,aboveskip=4pt]{caption}

\aclfinalcopy 

\title{GumDrop at the DISRPT2019 Shared Task: A Model Stacking Approach to Discourse Unit Segmentation and Connective Detection}

\author{
  Yue Yu \\ Computer Science \\ Georgetown University \And Yilun Zhu \\ Linguistics \\ Georgetown University \And Yang Liu \\ Linguistics \\ Georgetown University \AND Yan Liu \\ Analytics \\ Georgetown University \And Siyao Peng \\ Linguistics \\ Georgetown University \And Mackenzie Gong \\ CCT \\ Georgetown University \And Amir Zeldes \\ Linguistics \\ Georgetown University \AND
  {\tt \vspace{-1.5cm} \{yy476,yz565,yl879,yl1023,sp1184,mg1745,az364\}@georgetown.edu} \vspace{-1.5cm} \\} 

\date{}
\par\vspace{+20pt}\par
\begin{document}
\maketitle
\begin{abstract}
  In this paper we present GumDrop, Georgetown University's entry at the DISRPT 2019 Shared Task on automatic discourse unit segmentation and connective detection. Our approach relies on model stacking, creating a heterogeneous ensemble of classifiers, which feed into a metalearner for each final task. The system encompasses three trainable component stacks: one for sentence splitting, one for discourse unit segmentation and one for connective detection. The flexibility of each ensemble allows the system to generalize well to datasets of different sizes and with varying levels of homogeneity.
\end{abstract}

\section{Introduction}

Although discourse unit segmentation and connective detection are crucial for higher level shallow and deep discourse parsing tasks, recent years have seen more progress in work on the latter tasks than on predicting underlying segments, such as Elementary Discourse Units (EDUs). As the most recent overview on parsing in the framework of Rhetorical Structure Theory (RST, \citealt{MannThompson1988}) points out \cite[1322]{MoreyMullerAsher2017} ``all the parsers in our sample except [two] predict binary trees over manually segmented EDUs''. Recent discourse parsing papers (e.g. \citealt{LiLiChang2016}, \citealt{BraudCoavouxSoegaard2017}) have focused on complex discourse unit span accuracy above the level of EDUs, attachment accuracy, and relation classification accuracy. This is due in part to the difficulty in comparing systems when the underlying segmentation is not identical (see \citealt{MarcuEtAl1999}), but also because of a relatively stable SOA accuracy of EDU segmentation as evaluated on the largest RST corpus, the English RST Discourse Treebank (RST-DT, \citealt{CarlsonEtAl2003}), which already exceeded 90\% accuracy in 2010 \cite{HernaultPrendingerEtAl2010}. 

However, as recent work \cite{BraudLacroixSoegaard2017} has shown, performance on smaller or less homogeneous corpora than RST-DT, and especially in the absence of gold syntax trees (which are realistically unavailable at test time for practical applications), hovers around the mid 80s, making it problematic for full discourse parsing in practice. This is more critical for languages and domains in which relatively small datasets are available, making the application of generic neural models less promising. 

The DISRPT 2019 Shared Task aims to identify spans associated with discourse relations in data from three formalisms: RST \cite{MannThompson1988}, SDRT \cite{Asher1993} and PDTB \cite{PrasadWebberJoshi2014}. The targeted task varies actoss frameworks: Since RST and SDRT segment texts into spans covering the entire document, the corresponding task is to predict the starting point of new discourse units. In the PDTB framework, the basic locus identifying explicit discourse relations is the spans of discourse connectives which need to be identified among other words. In total, 15 corpora (10 from RST data, 3 from PDTB-style data, and 2 from SDRT) in 10 languages (Basque, Chinese, Dutch, English, French, German, Portuguese, Russian, Spanish, and Turkish) are used as the input data for the task. The heterogeneity of the frameworks, languages and even the size of the training datasets all render the shared task challenging:  training datasets range from the smallest Chinese RST corpus of 8,960 tokens to the largest English PDTB dataset of 1,061,222 tokens, and all datasets have some  different guidelines. In this paper, we therefore focus on creating an architecture that is not only tailored to resources like RST-DT, and takes into account the crucial importance of high accuracy sentence splitting for real-world data, generalizing well to different guidelines and datasets. 

Our system, called GumDrop, relies on model stacking \cite{WOLPERT1992241}, which has been successfully applied to a number of complex NLP problems (e.g. \citealt{ClarkManning2015}, \citealt{Friedrichs2017InfyNLPAS}). The system uses a range of different rule-based and machine learning approaches whose predictions are all fed to a `metalearner' or blender classifier, thus benefiting from both neural models where appropriate, and strong rule-based baselines coupled with simpler classifiers for smaller datasets. A further motivation for our model stacking approach is curricular: the system was developed as a graduate seminar project in the course LING-765 (Computational Discourse Modeling), and separating work into many sub-modules allowed each contributor to work on a separate sub-project, all of which are combined in the complete system as an ensemble. The system was built by six graduate students and the instructor, with each student focusing on one module (notwithstanding occasional collaborations) in two phases: work on a high-accuracy ensemble sentence splitter for the automatic parsing scenario (see Section \ref{sec:sentencer}), followed by the development of a discourse unit segmenter or connective detection module (Sections \ref{sec:segmentation} and \ref{sec:connective}).

\section{Previous Work}\label{sec:prev_work}

Following early work on rule-based segmenters (e.g. \citealt{Marcu2000}, \citealt{ThanhAbeysingheHuyck2004}), \citet{SoricutMarcu2003} used a simple probabilistic model conditioning on lexicalized constituent trees, by using the highest node above each word that has a right-hand sibling, as well as its children. Like our approach, this and subsequent work below  perform EDU segmentation as a token-wise binary classification task (boundary/no-boundary). In a more complex model, \citet{SporlederLapata2005} used a two-level stacked boosting classifier on syntactic chunks, POS tags, token and sentence lengths, and token positions within clauses, all of which are similar to or subsumed by some of our features below. They additionally used the list of English connectives from \citet{Knott1996} to identify connective tokens.

\citet{HernaultPrendingerEtAl2010} used an SVM model with features corresponding to token and POS trigrams at and preceding a potential segmentation point, as well as features encoding the lexical head of each token's parent phrase in a phrase structure syntax tree and the same features for the sibling node on the right. More recently, \citet{BraudLacroixSoegaard2017} used a bi-LSTM-CRF sequence labeling approach on dependency parses, with words, POS tags, dependency relations and the same features for each word's parent and grand-parent tokens, as well as the direction of attachment (left or right), achieving F-scores of .89 on segmenting RST-DT with parser-predicted syntax, and scores in the 80s, near or above previous SOA results, for a number of other corpora and languages.

By contrast, comparatively little work has approached discourse connective detection as a separate task, as it is usually employed as an intermediate step for predicting discourse relations.  \citet{pitler2009using} used a Max Entropy classifier using a set of syntactic features extracted from the gold standard Penn Treebank \cite{MarcusSantoriniMarcinkiewicz1993} parses of PDTB \cite{PrasadEtAl2008} articles, such as the highest node which dominates exactly and only the words in the connective, the category of the immediate parent of that phrase, and the syntactic category
of the sibling immediately to the left/right of the same phrase. \citet{patterson2013predicting} presented a logistic regression model trained on eight relation types extracted from PDTB, with features in three categories: \textit{Relation-level} features such as the connective signaling the relation, attribution status of the relation, and its relevance to financial information; \textit{Argument-level} features, capturing the size or complexity of each of its two arguments; and \textit{Discourse-level} features, which incorporate the dependencies between the relation in question and its neighboring relations in the text.

\citet{polepalli2012automatic} used SVM and CRF for identifying discourse connectives in biomedical texts. The Biomedical Discourse Relation Bank \cite{prasad2011biomedical} and PDTB were used for in-domain classifiers and novel domain adaptation respectively. Features included POS tags, the dependency label of tokens' immediate parents in a parse tree, and the POS of the left neighbor; domain-specific semantic features included several biomedical gene/species taggers, in addition to NER features predicted by ABNER (A Biomedical Named Entity Recognition). 

\section{GumDrop}

\begin{figure*}[hbt]
\centering
\includegraphics[width=\textwidth, trim={0.5cm 4.8cm 1.5cm 2.1cm}, clip]{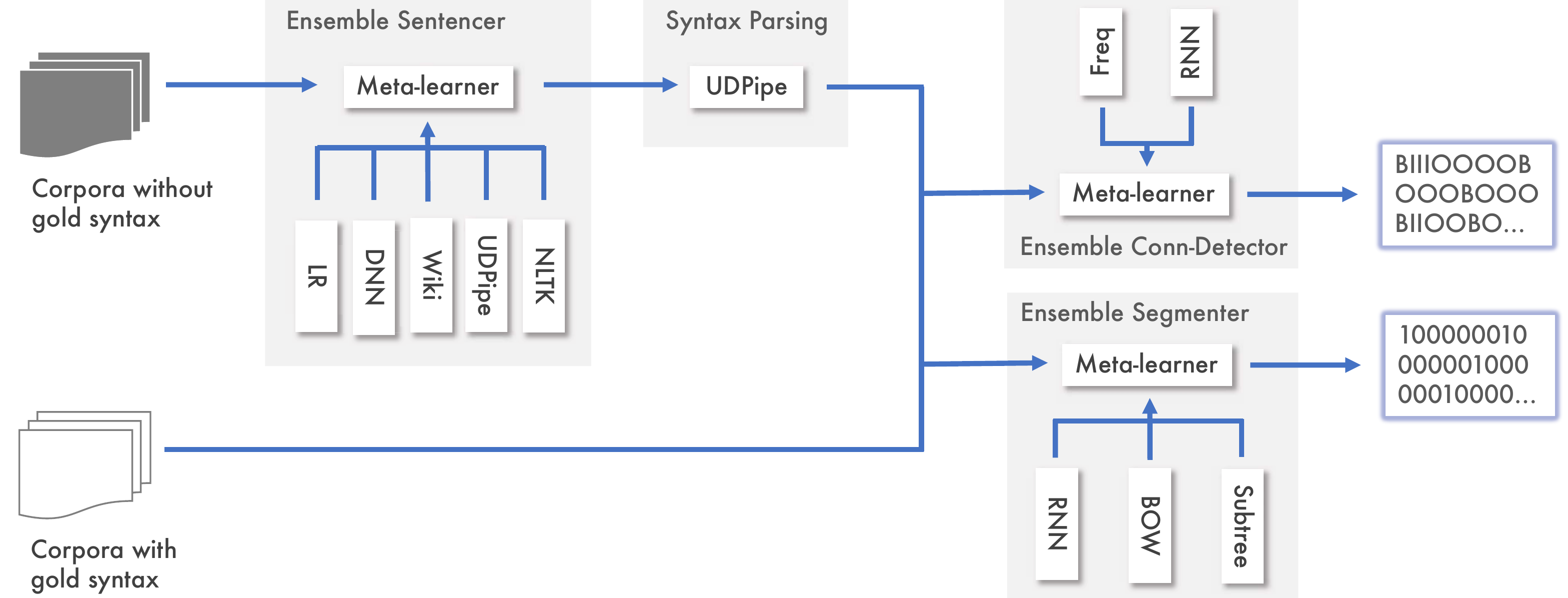}
\caption{System architecture. The raw text from corpora without gold syntax is first split into sentences by the ensemble sentencer. Sentences are then parsed using UDPipe. Corpora with predicted or gold syntax can then be utilized for discourse unit segmentation and connective detection.}
\label{fig:sysoverview}
\par\vspace{-10pt}\par
\end{figure*}

Our system is organized around three ensembles which implement model stacking.

\begin{enumerate}
  \setlength\itemsep{0.2em}
    \item A trainable sentencer ensemble which feeds an off-the-shelf dependency parser
    \item A discourse unit segmenter ensemble, operating on either gold or predicted sentences
    \item A connective detector ensemble, also using gold or predicted sentences
\end{enumerate}

Each module consists of several distinct sub-modules, as shown in Figure \ref{fig:sysoverview}. Predicted labels and probabilities from sub-modules, along with features for every token position are fed to a blender classifier, which outputs the final prediction for each token. By learning which modules perform better on which dataset, in which scenario (gold or predicted syntax) and in what linguistic environments, the ensemble remains robust at both tasks in both settings. 

Since the sub-modules and the ensembles are trained on the same training data, a crucial consideration is to avoid over-reliance on modules, which may occur if the metalearner learns about module reliability from data that the sub-modules have already seen. To counter this, we use 5-fold multitraining: each base module is trained five times, each time predicting labels for a disjoint held-out subset of the training data. These predictions are saved and fed to the ensemble as training data, thereby simulating the trained sub-modules' behavior when exposed to unseen data. At test time, live predictions are gathered from the sub-modules, whose reliability has been assessed via the prior unseen multitraining data.

\subsection{Features}

Table \ref{tab:feats} gives an overview of the features we extract from the shared task data, and the modules using those features for sentence splitting and EDU segmentation/connective detection. Features derived from syntax trees are not available for sentence splitting, though automatic POS tagging using the TreeTagger \cite{Schmid1994} was used as a feature for this task, due to its speed and good accuracy in the absence of sentence splits.


\begin{table*}[bt]
\begin{center}
\small
\begin{tabular}{c|c|c|c|c|c|c|c|c|c|c}
& \multicolumn{6}{c|}{\textbf{Sentence splitting}} &  
\multicolumn{4}{c}{\textbf{EDU/connective segmentation}} \\
\hline 
        Feature & \small{LR} & \small{NLTK} & \small{UDPipe} & \small{WikiSent} & \small{DNN} & \small{Meta} & \small{Subtree} & \small{RNN} & \small{BOW} & \small{Meta}  \\
\hline 
        word & n & y & y & y & y & top 100 & top 200 & y & top 200 & top 100 \\
        chars & f/l & n & y & n & n & n & n & y & n & n \\
        upos/xpos & y & n & n & n & n & y & y & y & y & y \\
        case & y & n & n & n & n & y & y & n & n & y \\
        \#char\_types & y & n & n & n & n & n & n & n & n & n \\
        tok\_len & y & n & n & n & n & y & y & n & n & y \\
        tok\_frq & y & n & n & n & n & n & n & n & n & n \\
        genre & n & n & n & n & n & y & y & y & n & y \\
        quot/paren & n & n & n & n & n & n & y & n & n & y \\
        sent\% & n & n & n & n & n & y & y & n & n & y \\
        deprel & -- & -- & -- & -- & -- & -- & y & y & n & y \\
        headdist  & -- & -- & -- & -- & -- & -- & y & bin & n & y \\
        depbracket & -- & -- & -- & -- & -- & -- & y & y & n & y \\
        children  & -- & -- & -- & -- & -- & -- & y & n & n & n \\
\hline
\end{tabular}
\end{center}
\caption{\label{tab:feats} Features for sentence splitting and EDU segmentation modules. }
\par\vspace{-10pt}\par
\end{table*}

    Most modules represent underlying words somehow, usually in a 3 or 5-gram window centered around a possible split point. An exception is the LR module, which uses only the first/last (f/l in Table \ref{tab:feats}) characters to prevent sparseness, but which also uses \texttt{\#char\_types} features, which give the count of digits, consonant, vowel and other characters per word. Modules with `top 200/100' use only the \textit{n} most frequent items in the data, and otherwise treat each word as its POS category. Neural modules (DNN, RNN) use 300 dimensional FastText \cite{bojanowski2017} word embeddings, and in the case of the RNN, character embeddings are also used. For Chinese in the LR module, we use the first/last byte in each word instead of actual characters.

The feature \texttt{genre} gives the genre, based on a substring extracted from document names, in corpora with multiple genres. The features \texttt{quot/paren} indicate, for each token, whether it is between quotation marks or parentheses, allowing modules to notice direct speech or uncompleted parentheses which often should not be split. The feature \texttt{sent\%} gives the quantile position of the current sentence in the document as a number between 0--1. This can be important for datasets in which position in the document interacts with segmentation behavior, such as abstracts in early portions of the academic genres in the Russian corpus, which often leave sentences unsegmented.

\begin{figure*}[h!bt]
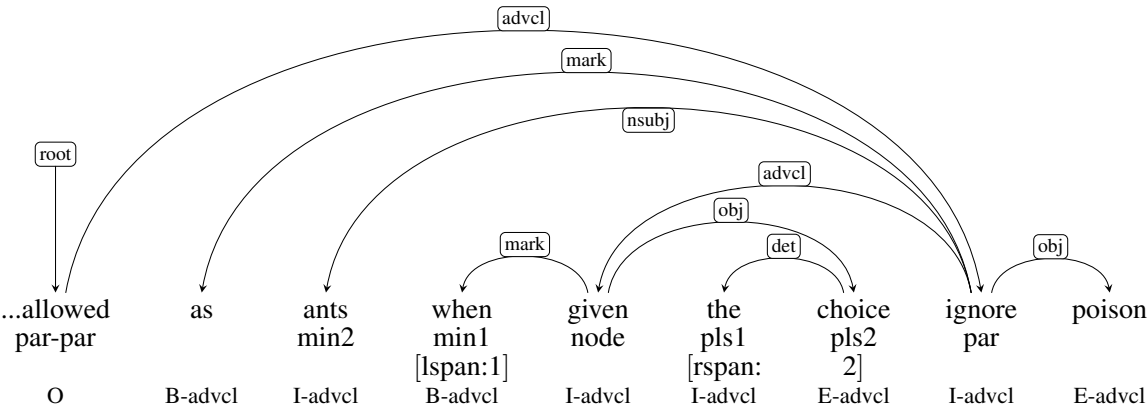

\centering
\resizebox{0.98 \textwidth}{!}{   
\begin{dependency}[arc edge, arc angle=80, label style={above}]
\begin{deptext}[column sep=.58cm]
...allowed \& as \& ants               \& when               \& given \& the \& choice \& ignore \& poison \\
par-par \&    \& min2               \& min1               \& node  \& pls1 \& pls2   \&  par      \& \\
  \&    \&                 \& $[$lspan:1$]$               \&   \& $[$rspan: \& 2$]$   \&        \& \\
  \small{O} \& \small{B-advcl} \& \small{I-advcl} \& \small{B-advcl} \& \small{I-advcl} \& \small{I-advcl} \& \small{E-advcl} \& \small{I-advcl} \& \small{E-advcl} \\
\end{deptext}
\deproot{1}{root}
\depedge{1}{8}{advcl}
\depedge{8}{5}{advcl}
\depedge{5}{4}{mark}
\depedge{5}{7}{obj}
\depedge{7}{6}{det}
\depedge{8}{2}{mark}
\depedge[label style={below}]{8}{3}{nsubj}
\depedge{8}{9}{obj}
\end{dependency}
}
\caption{Dependency features from a sentence fragment for a window surrounding `given' in SubtreeSegmenter.}\label{fig:children}
\par\vspace{-10pt}\par
\end{figure*}

The features \texttt{deprel}, \texttt{headdist} and \texttt{depbracket} are not available for sentence splitting, as they require dependency parses: they give the dependency relation, distance to the governing head token (negative/positive for left/right parents), and a BIEO (Begin/Inside/End/Out) encoded representation of the smallest relevant phrase boundaries covering each token for specific phrase types, headed by clausal functions such as `advcl', `xcomp' or `acl' (see Figure \ref{fig:children}). For the RNN, \texttt{headdist} is binned into 0, next-left/right, close-left/right (within 3 tokens) and far-left/right. The \texttt{children} feature set is unique to the Subtree module and is discussed below.

\subsection{Sentence Splitting}\label{sec:sentencer}
\textbf{DNN Sentencer} A simple Deep Neural Network classifier, using 300 dimensional word embeddings in a Multilayer Perceptron for tokens in a 5--9-gram window. Optimization on dev data determines the optimal window size for each dataset. Flexible window sizes enable the DNN model to remember the surrounding tokens in both small and large datasets. Starting and ending symbols (`$<$s$>$' and `$<$/s$>$') for each document guarantee the model can always predict the correct label when a new document starts.

\paragraph{Logistic Regression Sentencer}
The Logistic Regression (LR) Sentencer uses sklearn's \cite{PedregosaVaroquauxGramfortEtAl2011} LogisticRegressionCV implementation to predict sentence boundaries given a variety of character-level information. The beginning/ending characters (\textit{first/last letter}), auto-generated POS tags and character/frequency count representations (\textit{number of consonants/vowels/digits/other, token length, token frequency}) are applied to a sliding 5-gram window (categorical features are converted into 1-hot features). One advantage of the LR model is its reliability for smaller datasets where character-level features prevent sparseness (including the \textit{top 200} feature decreases performance).

\paragraph{Wiki-Based Sentencer}
The Wiki-Based Sentencer relies on the frequencies and ratios of paragraph-initial tokens extracted from Wikipedia articles obtained from Wikipedia database dumps for all languages.\footnote{ Traditional Chinese characters were converted into simplified Chinese to be consistent with shared task data.} The rationale is that even though we have no gold sentence splits for Wikipedia, if a token occurs paragraph-initial, then it must be sentence-initial. For each Wiki paragraph, we extract the first ``sentence'' based on text up to the first sentence final character (./?/!), and then the first word is obtained based on automatic tokenization. Though this approach is coarse, we are able to get a good approximation of frequently initial words thanks to the large data. The frequencies and ratios of tokens being sentence initial are recorded, and thresholds of frequency$>$10 and ratio $>$ 0.5 are set to collect the most relevant tokens. The main purpose of this module is to capture potential sentence split points such as headings, which are not followed by periods (e.g. \textit{Introduction} in English).  

\paragraph{UDPipe + NLTK}
Additionally, we used UDPipe and NLTK's freely available models as predictors for the ensemble. For Simplified Chinese, we retrained UPipe using data from the Chinese Treebank, not overlapping CDTB's shared task data.

\paragraph{EnsembleSentencer}
As a metalearner receiving input from the base-modules, we used tree-based algorithms selected via optimization on dev data, either RandomForest, ExtraTrees, GradientBoosting (using sklearn's implementation), or XGBoost \cite{ChenGuestrin2016}. In addition to the submodules' probability estimates, the metalearner was given access to token features in a trigram window, including word identity (for the top 100 items), POS tags, and orthographic case.

\subsection{Discourse Unit Segmentation}\label{sec:segmentation}

The feature space for segmentation is much larger than for sentence splitting, due to availability of syntactic features (cf. Table \ref{tab:feats}). Additionally, as usefulness of features varies across datasets (for example, some lanaguage use only the UPOS column, or UPOS is trivially predictable from XPOS), we performed automatic variable filtering per dataset for both the Subtree and the Ensemble module below. We removed all categorical variables with a Theil's U value of implication above .98 (meaning some feature A is predictable based on some feature B), and for numerical variables, based on Pearson's r$>$0.95. 

\paragraph{SubtreeSegmenter}
This module focuses on dependency subgraphs, looking at a trigram around the potential split point. In addition to word, orthographic case, POS, and deprel features from Table \ref{tab:feats}, the module uses a \texttt{children} feature set, extracting information for the node token, neighbors, parent and grandparent, including:

\begin{itemize}
  \setlength\itemsep{0.05em}
    \item their labels and depth (rank) in the tree
    \item labels of closest/farthest L/R children
    \item left/right span length and clause BIOE
    \item whether L/R neighbors share their parent
\end{itemize}

The features are illustrated in Figure \ref{fig:children}. If we consider a split at the node word `given', we collect features for two tokens in each direction, the parent (`ignore') and grandparent (`allowed'). The left span of children of `given' is 1 token long, and the right 2 tokens long. We additionally collect for each of these tokens whether they have the same parent as their neighbor to the right/left (e.g. `ants' has the same parent as `as'), as well as the nearest and farthest dependency label on descendents to each side of the node (here, \textit{mark} for both closest and farthest left child of `given', and \textit{det} (closest) and \textit{obj} (farthest) on the right. The BIOE bracket feature is a flattened `chunk' feature indicating clauses opening and closing (\textsc{B-advcl}, etc.) These features give a good approximation of the window's syntactic context, since even if the split point is nested deeper than a relevant clausal function, discrepancies in neighbors' dependency features, and distances implied by left/right spans along with dependency functions allow the reconstruction of pertinent subtree environments for EDU segmentation. The feature count varied between 86--119 (for rus.rst.rrt and eng.sdrt.stac respectively), due to automatic feature selection.

\paragraph{BOWCounter}
Rather than predicting exact split points, the BOWCounter attempts to predict the number of segments in each sentence, using a Ridge regressor with regularization optimized via cross-validation. The module uses the top 200 most frequent words as well as POS tags in a bag of words model and predicts a float which is fed directly to the ensemble. This allows the module to express confidence, rather than an integer prediction. We note that this module is also capable of correctly predicting 0 segmentation points in a sentence (most frequent in the Russian data).

\paragraph{RNNSegmenter}
To benefit from the predictive power of neural sequence models and word embeddings with good coverage for OOV items, we used NCRF++ \cite{yang2018ncrf}, a bi-LSTM/CNN-CRF sequence labeling framework. Features included Glove word embeddings for English \cite{PenningtonSocherManning2014} and FastText embeddings \cite{bojanowski2017} for other languages, trainable character embeddings, as well as the features in Table \ref{tab:feats}, such as POS tags, dependency labels, binned distance to parent, genre, and BIEO dependency brackets, all encoded as dense embeddings. We optimized models for each dataset, including using CNN or LSTM encoding for character and word embeddings.  

\paragraph{Ensemble Segmenter}
For the metalearner we used XGBoost, which showed high accuracy across dataset sizes. The ensemble was trained on serialized multitraining data, produced by training base-learners on 80\% of the data and predicting labels for each 20\% of the training data separately. At test time, the metalearner then receives live predictions from the sub-modules, whose reliability has been assessed using the multitraining data. In addition to base module predictions, the metalearner is given access to the most frequent lexemes, POS tags, dependency labels, genre, sentence length, and dependency brackets, in a trigram window.

\begin{table*}[h!]
\begin{center}
\small
\begin{tabular}{c|c|c|c|c|c|c|c|c|c|c|c|c}
& \multicolumn{3}{c|}{\textbf{Baseline (./!/?)}} &
\multicolumn{3}{|c}{\textbf{NLTK}} &
\multicolumn{3}{|c}{\textbf{LR}} &
\multicolumn{3}{|c}{\textbf{GumDrop}} \\
\hline 
corpus & P & R & F & P & R & F & P & R & F & P & R & F \\
\hline
deu.rst.pcc & 1.00 & .864 & .927 & 1.00 & .864 & .927 & .995 & .953 & .974 & .986 & .986 & \textbf{.986} \\
eng.pdtb.pdtb & .921 & .916 & .918 & .899 & .863 & .880 & .891 & .970 & .929 & .963 & .948 & \textbf{.955} \\
eng.rst.gum & .956 & .810 & .877 & .943 & .807 & .870 & .935 & .885 & .909 & .977 & .874 & \textbf{.923} \\
eng.rst.rstdt & .901 & .926 & .913 & .883 & .900 & .891 & .897 & .991 & .942 & .963 & .946 & \textbf{.954} \\
eng.sdrt.stac & .961 & .290 & .446 & .990 & .283 & .440 & .805 & .661 & .726 & .850 & .767 & \textbf{.806} \\
eus.rst.ert & .964 & 1.00 & .982 & .945 & .972 & .958 & 1.00 & 1.00 & \textbf{1.00} & 1.00 & .997 & .998 \\
fra.sdrt.annodis & .970 & .910 & .939 & .965 & .910 & .937 & .957 & .943 & .950 & .985 & .905 & \textbf{.943} \\
nld.rst.nldt & .991 & .919 & .954 & .983 & .919 & .950 & .951 & .931 & .941 & .980 & .964 & \textbf{.972} \\
por.rst.cstn & .984 & .992 & \textbf{.988} & .967 & .967 & .967 & .984 & .992 & \textbf{.988} & .984 & .984 & \textbf{.988} \\
rus.rst.rrt & .867 & .938 & .901 & .737 & .927 & .821 & .948 & .980 & \textbf{.964} & .952 & .972 & .962 \\
spa.rst.rststb & .912 & .851 & .881 & .938 & .845 & .889 & .996 & .934 & \textbf{.964} & .993 & .934 & .963 \\
spa.rst.sctb & .860 & .920 & .889 & .852 & .920 & .885 & .889 & .960 & \textbf{.923} & .857 & .960 & .906 \\
tur.pdtb.tdb & .962 & .922 & .942 & .799 & .099 & 176 & .979 & .979 & .979 & .983 & .984 & \textbf{.983} \\
zho.pdtb.cdtb & .959 & .866 & .910 & .-- & .-- & .-- & .954 & .975 & .965 & .980 & .975 & \textbf{.978} \\
zho.rst.sctb & .879 & .826 & .852 & .-- & .-- & .-- & 1.00 & .811 & \textbf{.895} & .991 & .795 & .882 \\
\hline
\textbf{mean} & .939 & .863 & .888 & .915 & .790 & .815 & .945 & .931 & .937 & .963 & .933 & \textbf{.947} \\
\textbf{std} & .046 & 167 & 128 & .079 & .273 & .235 & .055 & .089 & .065 & .046 & .070 & \textbf{.050}

\end{tabular}
\end{center}
\caption{\label{tab:sent} GumDrop sentence splitting performance. }
\par\vspace{-15pt}\par
\end{table*}

\subsection{Connective Detection}\label{sec:connective}

\paragraph{Frequency-based Connective Detector}
This module outputs the ratios at which sequences of lexical items have been seen as connectives in training data, establishing an intelligent `lookup' strategy for the connective detection task. Since connectives can be either a single \textsc{B-Conn} or a \textsc{B-Conn} followed by several \textsc{I-Conn}s, we recover counts for each attested connective token sequence up to 5 tokens. For test data, the module reports the longest possible connective sequence containing a token and the ratio at which it is known to be a connective, as well as the training frequency of each item. Rather than select a cutoff ratio for positive prediction, we allow the ensemble to use the ratio and frequency dynamically as features.

\paragraph{RNN Connective Detector}
This module is architecturally identical to the RNN EDU segmenter, but since connective labels are non-binary and may form spans, it classifies sequences of tokens with predicted connective types (i.e. \textsc{B-Conn}, \textsc{I-Conn} or not a connective). Rather than predicted labels, the system reports probabilities with which each label is suspected to apply to tokens, based on the top 5 optimal paths as ranked by the CRF layer of NCRF$++$'s output.

\paragraph{Ensemble Connective Detector}
The connective ensemble is analogous to the segmenter ensemble, and relies on a Random Forest classifier fed the predicted labels and probabilities from base connective detectors, as well as the same features fed to the segmenter ensemble above.

\section{Results}

\paragraph{Sentence Splitting}
Although not part of the shared task, we report results for our EnsembleSentencer and LR module (best sub-module on average) next to a punctuation-based baseline (split on `.', `!', `?' and Chinese equivalents) and NLTK's \cite{BirdLoperKlein2009} sentence tokenizer (except for Chinese, which is not supported). Since most sentence boundaries are also EDU boundaries, this task is critical, and Table \ref{tab:sent} shows the gains brought by using the ensemble. GumDrop's performance is generally much higher than both baselines, except for the Portuguese corpus, in which both the system and the baseline make exactly 2 precision errors and one recall error, leading to an almost perfect tied score of 0.988. Somewhat surprisingly, NLTK performs worse on average than the conservative strategy of using sentence final punctuation. The LR module is usually slightly worse than the ensemble, but occasionally wins by a small margin.

\begin{table*}[h]
\begin{center}
\small
\begin{tabular}{c|c|c|c|c|c|c|c|c|c|c|c|c}
\textbf{Gold syntax} & \multicolumn{3}{c|}{\textbf{Baseline}} &  
\multicolumn{3}{|c}{\textbf{Subtree}} &
\multicolumn{3}{|c}{\textbf{RNN}} &
\multicolumn{3}{|c}{\textbf{GumDrop}} \\
\hline 
corpus & P & R & F & P & R & F & P & R & F & P & R & F \\

\hline
deu.rst.pcc & 1.0 & .724 & .840 & .960 & .891 & \textbf{.924} & .892 & .871 & .881 & .933 & .905 & .919 \\
eng.rst.gum & 1.0 & .740 & .850 & .974 & .888 & .929 & .950 & .877 & .912 & .965 & .908 & \textcolor{purple}{\textbf{.935}} \\
eng.rst.rstdt & 1.0 & .396 & .567 & .951 & .945 & .948 & .932 & .945 & .939 & .949 & .965 & \textbf{.957} \\
eng.sdrt.stac & .999 & .876 & .933 & .968 & .930 & .949 & .946 & .971 & \textcolor{purple}{\textbf{.958}} & .953 & .954 & \textcolor{purple}{.953} \\
eus.rst.ert & .981 & .530 & .688 & .890 & .707 & .788 & .889 & .754 & .816 & .909 & .740 & \textbf{.816} \\
fra.sdrt.annodis & 1.0 & .310 & .474 & .943 & .854 & .897 & .894 & .903 & .898 & .944 & .865 & \textbf{.903} \\
nld.rst.nldt & 1.0 & .721 & .838 & .979 & .927 & .952 & .933 & .892 & .912 & .964 & .945 & \textcolor{purple}{\textbf{.954}} \\
por.rst.cstn & .878 & .435 & .582 & .911 & .827 & .867 & .815 & .903 & .857 & .918 & .899 & \textbf{.908} \\
rus.rst.rrt & .760 & .490 & .596 & .809 & .745 & .775 & .821 & .710 & .761 & .835 & .755 & \textbf{.793} \\
spa.rst.rststb & .974 & .647 & .777 & .921 & .792 & .851 & .759 & .855 & .804 & .890 & .818 & \textbf{.853} \\
spa.rst.sctb & .970 & .577 & .724 & .938 & .631 & .754 & .901 & .649 & .754 & .898 & .679 & \textbf{.773} \\
zho.rst.sctb & .924 & .726 & .813 & .880 & .744 & .806 & .843 & .768 & .804 & .810 & .810 & \textbf{.810} \\
\hline
mean & .957 & .598 & .724 & .927 & .823 & .870 & .881 & .841 & .858 & .914 & .853 & \textbf{.881} \\

\hline\hline

\textbf{Pred syntax} & \multicolumn{3}{c|}{\textbf{Baseline}} &  
\multicolumn{3}{|c}{\textbf{Subtree}} &
\multicolumn{3}{|c}{\textbf{RNN}} &
\multicolumn{3}{|c}{\textbf{GumDrop}} \\
\hline 
corpus & P & R & F & P & R & F & P & R & F & P & R & F \\
\hline
deu.rst.pcc & 1.0 & .626 & .770 & .924 & .867 & .895 & .876 & .867 & .872 & .920 & .898 & \textbf{.909} \\
eng.rst.gum & .956 & .599 & .737 & .948 & .777 & .854 & .910 & .805 & .854 & .940 & .772 & \textbf{.848} \\
eng.rst.rstdt & .906 & .368 & .524 & .916 & .871 & .893 & .883 & .911 & .897 & .896 & .914 & \textbf{.905} \\
eng.sdrt.stac & .956 & .253 & .401 & .849 & .767 & .806 & .819 & .814 & .817 & .842 & .775 & \textbf{.807} \\
eus.rst.ert & .970 & .543 & .696 & .917 & .705 & .797 & .877 & .747 & .807 & .901 & .734 & \textbf{.809} \\
fra.sdrt.annodis & .980 & .285 & .442 & .938 & .824 & .877 & .892 & .915 & \textbf{.903} & .945 & .853 & .896 \\
nld.rst.nldt & .991 & .663 & .794 & .951 & .849 & .897 & .938 & .835 & .883 & .947 & .884 & \textbf{.915} \\
por.rst.cstn & .879 & .440 & .586 & .935 & .867 & \textbf{.900} & .788 & .883 & .833 & .930 & .851 & .888 \\
rus.rst.rrt & .664 & .463 & .545 & .825 & .717 & .767 & .813 & .731 & .770 & .821 & .748 & \textbf{.783} \\
spa.rst.rststb & .912 & .566 & .698 & .934 & .772 & \textbf{.845} & .820 & .871 & \textbf{.845} & .875 & .798 & .835 \\
spa.rst.sctb & .888 & .565 & .691 & .870 & .637 & .735 & .813 & .595 & .687 & .853 & .655 & \textbf{.741} \\
zho.rst.sctb & .798 & .589 & .678 & .806 & .643 & .715 & .803 & .607 & .692 & .770 & .696 & \textcolor{purple}{\textbf{.731}} \\
\hline
mean & .908 & .497 & .630 & .901 & .775 & .832 & .853 & .798 & .822 & .887 & .798 & \textbf{.839} \\

\end{tabular}
\end{center}
\caption{\label{tab:seg} Subtree, RNN and full GumDrop discourse unit segmentation performance.}
\end{table*}

\paragraph{Discourse Unit Segmentation}
Table \ref{tab:seg} gives scores for both the predicted and gold syntax scenarios. In order to illustrate the quality of the submodules, we also include scores for Subtree (the best non-neural model) and the RNN (best neural model), next to the ensemble. The baseline is provided by assuming EDUs overlap exactly with sentence boundaries.

Overall the results compare favorably with previous work and exceed the previously reported state of the art for the benchmark RST-DT dataset, in both gold and predicted syntax (to the best of our knowledge, 93.7 and 89.5 respectively). At the same time, the ensemble offers good performance across dataset sizes and genres: scores are high on all English datasets, covering a range of genres, including gold STAC (chat data), as well as on some of the smaller datasets, such as Dutch, French and German (only 17K, 22K and 26K training tokens each). Performance is worse on the SCTB corpora and Russian, which may be due to low-quality parses in the gold scenario, and some inconsistencies, especially in the Russian data, where academic abstracts and bibliographies were sometimes segmented and sometimes not. Comparing the ensemble to the RNN or subtree modules individually shows that although they each offer rather strong performance, the ensemble outperforms them for all datasets, except German, where Subtree outperforms it by a small margin, and STAC, where the RNN is slightly better, both showing just half a point of improvement.

For automatically parsed data, the table clearly shows that \textit{eng.rst.stac}, \textit{eng.rst.gum} and \textit{zho.rst.sctb} are the most problematic, in the first case since chat turns must be segmented automatically into sentences. This indicates that a trustworthy sentencer is crucial for discourse unit segmentation and thus very useful for this shared task. Here the EnsembleSentencer brings results up considerably from the punctuation based baseline. The ensemble achieves top performance for most datasets and on average, but the RNN performs better on French, Subtree on Portuguese, and both are tied for Spanish RSTSTB.

\paragraph{Connective Detection}
Results for connective detection are shown in Table \ref{tab:conn}. As a baseline, we consider assigning each word in the test data a connective label if and only if it is attested exclusively as a connective in the training set (case-sensitive). As the results show, the baseline has low recall but high precision, correlated with the size of the corpus (as exhaustivity of exclusive connective words increases with corpus size). 

\begin{table*}[hbt]
\begin{center}
\small
\begin{tabular}{c|c|c|c|c|c|c|c|c|c|c|c|c}
\textbf{Gold syntax} & \multicolumn{3}{c|}{\textbf{Baseline}} &  
\multicolumn{3}{|c}{\textbf{Freq}} &
\multicolumn{3}{|c}{\textbf{RNN}} &
\multicolumn{3}{|c}{\textbf{GumDrop}} \\
\hline 
corpus & P & R & F & P & R & F & P & R & F & P & R & F \\

\hline
eng.pdtb.pdtb &.964 &.022 &.044 &.836 &.578 &.683 &.859 &.871 &.865 &.879 &.888 &	\textbf{.884} \\
tur.pdtb.tdb &.333 &.001 &.002 &.786 &.355 &.489 &.759 &.820 &.788 &.766 &.816 &	\textcolor{purple}{\textbf{.790}} \\
zho.pdtb.cdtb &.851 &.259 &.397 &.715 &.618 &.663 &.726 &.628 &.674 &.813 &.702 &	\textbf{.754} \\
\hline
mean &.716 &.094 &.148 &.779 &.517 &.612 &.781 &.773 &.776 &.819 &.802 &	\textcolor{purple}{\textbf{.809}} \\

\hline\hline

\textbf{Pred syntax} & \multicolumn{3}{c|}{\textbf{Baseline}} &  
\multicolumn{3}{|c}{\textbf{Freq}} &
\multicolumn{3}{|c}{\textbf{RNN}} &
\multicolumn{3}{|c}{\textbf{GumDrop}} \\
\hline
corpus & P & R & F & P & R & F & P & R & F & P & R & F \\
\hline
eng.pdtb.pdtb &.964 &.022 &.044 &.836 &.578 &.683 &.811 &.798 &.805 &.846 &.828 &	\textbf{.837} \\
tur.pdtb.tdb &.333 &.001 &.002 &.786 &.355 &.489 &.761 &.821 &.790 &.768 &.817 &	\textbf{.792} \\
zho.pdtb.cdtb &.851 &.259 &.397 &.715 &.618 &.663 &.705 &.590 &.642 &.806 &.673 &	\textbf{.734} \\
\hline
mean &.716 &.094 &.148 &.779 &.517 &.612 &.759 &.736 &.746 &.806 &.773 & \textbf{.788} \\

\end{tabular}
\end{center}
\caption{\label{tab:conn} Connective detection performance.}
\end{table*}

The frequency-based connective detector gives a reasonable result with a rather simple strategy, using a threshold of 0.5 as the connective detection ratio. More importantly, it is useful as input for the ensemble that outperforms the sequence labeling RNN by itself on every dataset. We suspect at least two factors are responsible for this improvement: firstly, the imbalanced nature of connective annotations (the vast majority of words are not connectives) means that the RNN achieves over 99\% classification accuracy, and may have difficulty generalizing to rare but reliable connectives. Secondly, the RNN may overfit spurious features in the training data, to which the frequency detector is not susceptible. Coupled with the resistance of tree ensembles to overfitting and imbalanced problems, the ensemble is able to give a better solution to the task.

\section{Error Analysis}
\subsection{EDU Segmenter}\label{subsec:error_segmentation}
In both gold and predicted syntax scenarios, the RST corpora in Russian, Spanish and Chinese (rst.rus.rrt, spa.rst.sctb and zho.rst.sctb) achieve the lowest F-scores on this task. Leaving the sentencer performance aside, this error analysis for EDU segmentation will mainly focus on the gold syntax scenario of these three corpora.

\paragraph{Coordinating Conjunctions (CCONJ)}
Only particular types of coordinated structure consist of two discourse units in different corpora, e.g. VP coordination, or each coordinate predicate having its own subject, etc. For example, in eng.rst.gum, two coordinated verb phrases (\textit{[John is athletic but hates hiking]} are annotated as one discourse unit whereas \textit{[John is athletic] [but he hates hiking]} is divided into two units since both coordinates have their own subjects. Additionally, if one coordinate VP has a dependent adverbial clause, multiple units are annotated. However, even with dependency features included in GumDrop, precision and recall errors happen with different coordinating conjunctions. These include \textit{and}, \textit{or} in English, \textit{y} (`and'), \textit{o} (`or') in Spanish, and \textit{i} (`and'), \textit{a} (`but'), \textit{ili} (`or') in Russian.

\paragraph{Subordinating Conjunctions (SCONJ)}
GumDrop sometimes fails when there is an ambiguity between adpositions and subordinating conjunctions. Words that can function as both cause problems for segmentation since subordinate clauses are discourse units but adpositional phrases are not in most datasets. Ambiguous tokens include \textit{to}, \textit{by}, \textit{after}, \textit{before} in English, \textit{en} (`in'), \textit{de} (`of'), \textit{con} (`with'), \textit{por} (`by') in Spanish, as well as \textit{zai} (`at') in Chinese. 

\begin{figure}[h!bt]
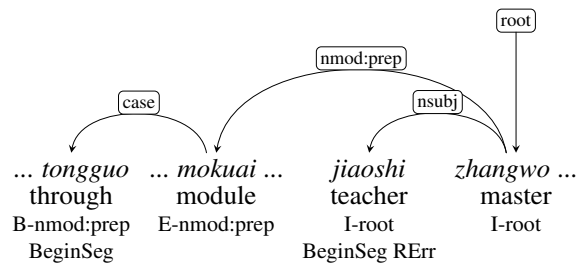

\centering
\resizebox{0.5 \textwidth}{!}{   

\begin{dependency}[arc edge, arc angle=80, label style={above}]
\begin{deptext}[column sep=.10cm]
...  \textit{tongguo} \& ... \textit{mokuai} ... \& \textit{jiaoshi} \& \textit{zhangwo} ... \\
 through \&  module   \& teacher \& master  \\
\small{B-nmod:prep} \& \small{E-nmod:prep} \& \small{I-root} \& \small{I-root} \\
\small{BeginSeg} \& \& \small{BeginSeg RErr} \& \\
\end{deptext}
\deproot{4}{root}
\depedge{4}{2}{nmod:prep}
\depedge{2}{1}{case}
\depedge{4}{3}{nsubj}
\end{dependency}
}
\caption{Example of a main clause preceded by a subordinate clause in zho.rst.sctb that causes a Recall Error (RErr) on the second instance of BeginSeg.}\label{fig:subordinate_zho}
\par\vspace{-10pt}\par
\end{figure}

Classifying the boundary of subordinate clauses is another problem. The \textit{depbracket} feature can identify the beginning of a subordinate clause when the main clause precedes it. However, when they are in reverse order as in Figure \ref{fig:subordinate_zho}, GumDrop fails to identify the beginning of the second discourse unit possibly due to the absence of a second B-feature at \textit{jiaoshi}. 

\paragraph{Enumerations and Listings}
In rus.rst.rrt, the special combination of a number, a backslash and a period, e.g. \textit{1\textbackslash .} , \textit{2\textbackslash .} etc., is used for enumeration. However, their dependency labels vary: \textit{root}, \textit{flat},  \textit{nmod}  etc. Due to the instability of the labels, these tokens may result in recall errors, suggesting possibile improvements via parser postprocessing. Similar errors also occur with \textit{1}, \textit{2} in Spanish and variants of hyphens/dashes in Russian.



\subsection{Connective Detection}\label{subsec:error_connective}

\paragraph{Co-occurring Connective Spans}
Unlike EDU segmentation, where only splits are marked, connectives are spans that consist of a mandatory B-Conn and possible I-Conn labels. However, in Chinese, it is possible for a a connective to consist of discontinuous spans. In (1), both \textit{zai} `at' and the localizer \textit{zhong}, are connectives and are required to co-occur in the context. However, the system fails to capture the relationship between them.
\\\\
\begin{tabular}{lllll}
    (1) & \textbf{zai} &  cunmin & zizhi & \textbf{zhong} ... \\
   & \texttt{P:}at & villager & autonomy & \texttt{LC}:in \\
   & B-Conn &  & & B-Conn \\
   & \multicolumn{4}{l}{`Under the autonomy of villagers...'} 
\end{tabular}


\paragraph{Syntactic Inversions} 
Syntactic inversion as a connective is also problematic since no content words are involved: For instance, though the system is able to identify B-Conn in both (2) and (3), it is hard to determine whether content words, such as the verbs (\textit{fueling} and \textit{yinrenzhumu}), belong to the connective span or not. The model can be potentially improved by handling these using dependency features. 
\\\\
\begin{tabular}{lllll}
    (2) & \textbf{Further} & \textbf{fueling} & the belief that ...  \\
   & B-Conn  & I-Conn & 
\end{tabular}
\\
\begin{tabular}{lllll}
    (3) & ... \textbf{geng} & \textbf{yinrenzhumude} & de & shi ... \\
    & more  & striking & \texttt{DE} & \texttt{COP} \\
    & B-Conn & I-Conn & & \\
   & \multicolumn{4}{l}{`the more striking thing is that ...'} 
\end{tabular}


\section{Conclusion and Future Work}

A main lesson learned from the present work has been that while RNNs perform well on large and consistent datasets, such as RST-DT, they are not as robust when dealing with smaller datasets. This was especially apparent in the predicted syntax scenario, where decision tree ensembles outperformed the RNN on multiple datasets. At the same time, the model stacking approach offers the advantage of not having to choose between neural and tree-based models, by letting a metalearner learn who to believe and when.

Although we hope these results on the shared task dataset represent progress on discourse unit segmentation and connective detection, we would also like to point out that high accuracy (95\% or better) is still out of reach, and especially so for languages with fewer resources and in the realistic `no gold syntax' scenario. Additionally, the architecture used in this paper trades improvements in accuracy for a higher level of complexity, including complex training regimes due to multitraining and a variety of supporting libraries. In future work, we plan to integrate a simplified version of the system into tools that are easier to distribute. In particular, we aim to integrate automatic segmentation facilities into rstWeb \cite{Zeldes2016}, an open source RST editor interface, so that end users can more easily benefit from system predictions.

\bibliography{naaclhlt2019}
\bibliographystyle{acl_natbib}

\end{document}